\documentclass[10pt,twocolumn,letterpaper]{article}

\usepackage{cvpr}              

\usepackage{graphicx}
\usepackage{amsmath}
\usepackage{amssymb}
\usepackage{booktabs}
\usepackage{listings}
\usepackage{mathtools}
\usepackage{cite}
\usepackage{amsmath,amssymb,amsfonts}
\usepackage{stfloats}
\usepackage{algorithmic}
\usepackage{graphicx}
\usepackage{textcomp}
\usepackage{xcolor}
\usepackage{multirow}
\usepackage{makecell}
\usepackage{pifont}
\usepackage{xspace}
\usepackage{colortbl}

\usepackage{tabularx}
\usepackage{subcaption}
\usepackage{caption}
\usepackage[misc]{ifsym}
\usepackage[accsupp]{axessibility}

\newcommand{\cmark}{\ding{51}}
\newcommand{\xmark}{\ding{55}}

\usepackage[pagebackref,breaklinks,colorlinks]{hyperref}

\makeatletter
\def\UrlAlphabet{%
      \do\a\do\b\do\c\do\d\do\e\do\f\do\g\do\h\do\i\do\j%
      \do\k\do\l\do\m\do\n\do\o\do\p\do\q\do\r\do\s\do\t%
      \do\u\do\v\do\w\do\x\do\y\do\z\do\A\do\B\do\C\do\D%
      \do\E\do\F\do\G\do\H\do\I\do\J\do\K\do\L\do\M\do\N%
      \do\O\do\P\do\Q\do\R\do\S\do\T\do\U\do\V\do\W\do\X%
      \do\Y\do\Z}
\def\UrlDigits{\do\1\do\2\do\3\do\4\do\5\do\6\do\7\do\8\do\9\do\0}
\g@addto@macro{\UrlBreaks}{\UrlOrds}
\g@addto@macro{\UrlBreaks}{\UrlAlphabet}
\g@addto@macro{\UrlBreaks}{\UrlDigits}
\makeatother

\usepackage[capitalize]{cleveref}
\crefname{section}{Sec.}{Secs.}
\Crefname{section}{Section}{Sections}
\Crefname{table}{Table}{Tables}
\crefname{table}{Tab.}{Tabs.}

\newcommand\blfootnote[1]{%
  \begingroup
  \renewcommand\thefootnote{}\footnote{#1}%
  \addtocounter{footnote}{-1}%
  \endgroup
}

\def\cvprPaperID{2452} 
\def\confName{CVPR}
\def\confYear{2023}

\begin{document}

\title{STMixer: A One-Stage Sparse Action Detector}

\author{Tao Wu\textsuperscript{1,*} \quad \quad Mengqi Cao\textsuperscript{1,*} \quad \quad Ziteng Gao\textsuperscript{1} \quad \quad Gangshan Wu\textsuperscript{1} \quad \quad Limin Wang\textsuperscript{1,2,~\Letter}\\
$^1$State Key Laboratory for Novel Software Technology, Nanjing University \quad $^2$Shanghai AI Lab \\
{\tt\small  \{wt,mg20370004\}@smail.nju.edu.cn, gzt@outlook.com, \{gswu, lmwang\}@nju.edu.cn}}
\maketitle

\begin{abstract}
Traditional video action detectors typically adopt the two-stage pipeline, where a person detector is first employed to generate actor boxes and then 3D RoIAlign is used to extract actor-specific features for classification. This detection paradigm requires multi-stage training and inference, and cannot capture context information outside the bounding box. Recently, a few query-based action detectors are proposed to predict action instances in an end-to-end manner. However, they still lack adaptability in feature sampling and decoding, thus suffering from the issues of inferior performance or slower convergence. In this paper, we propose a new one-stage sparse action detector, termed STMixer. STMixer is based on two core designs. First, we present a query-based adaptive feature sampling module, which endows our STMixer with the flexibility of mining a set of discriminative features from the entire spatiotemporal domain. Second, we devise a dual-branch feature mixing module, which allows our STMixer to dynamically attend to and mix video features along the spatial and the temporal dimension respectively for better feature decoding. Coupling these two designs with a video backbone yields an efficient end-to-end action detector. Without bells and whistles, our STMixer obtains the state-of-the-art results on the datasets of AVA, UCF101-24, and JHMDB.
\end{abstract}
\blfootnote{*: Equal contribution. \Letter: Corresponding author.}

\section{Introduction}

Video action detection~\cite{fat,hcstal,online,learningtrack,hfcn,multitwo,stagn} is an important problem in video understanding, which aims to recognize all action instances present in a video and also localize them in both space and time. It has drawn a significant amount of research attention, due to its wide applications in many areas like security and sports analysis. 

Since the proposal of large-scale action detection benchmarks~\cite{ava, multisports}, action detection has made remarkable progress. This progress is partially due to the advances of video representation learning such as video convolution neural networks~\cite{c3d,s3d,i3d,r2p1d,slowfast,non-local,csn} and video transformers~\cite{mvit,timesformer,vivit,videoswin,vidtr,vmae,VideoMAEv2}.

\begin{figure}[t]
    \centering
    \includegraphics[width=0.4\textwidth]{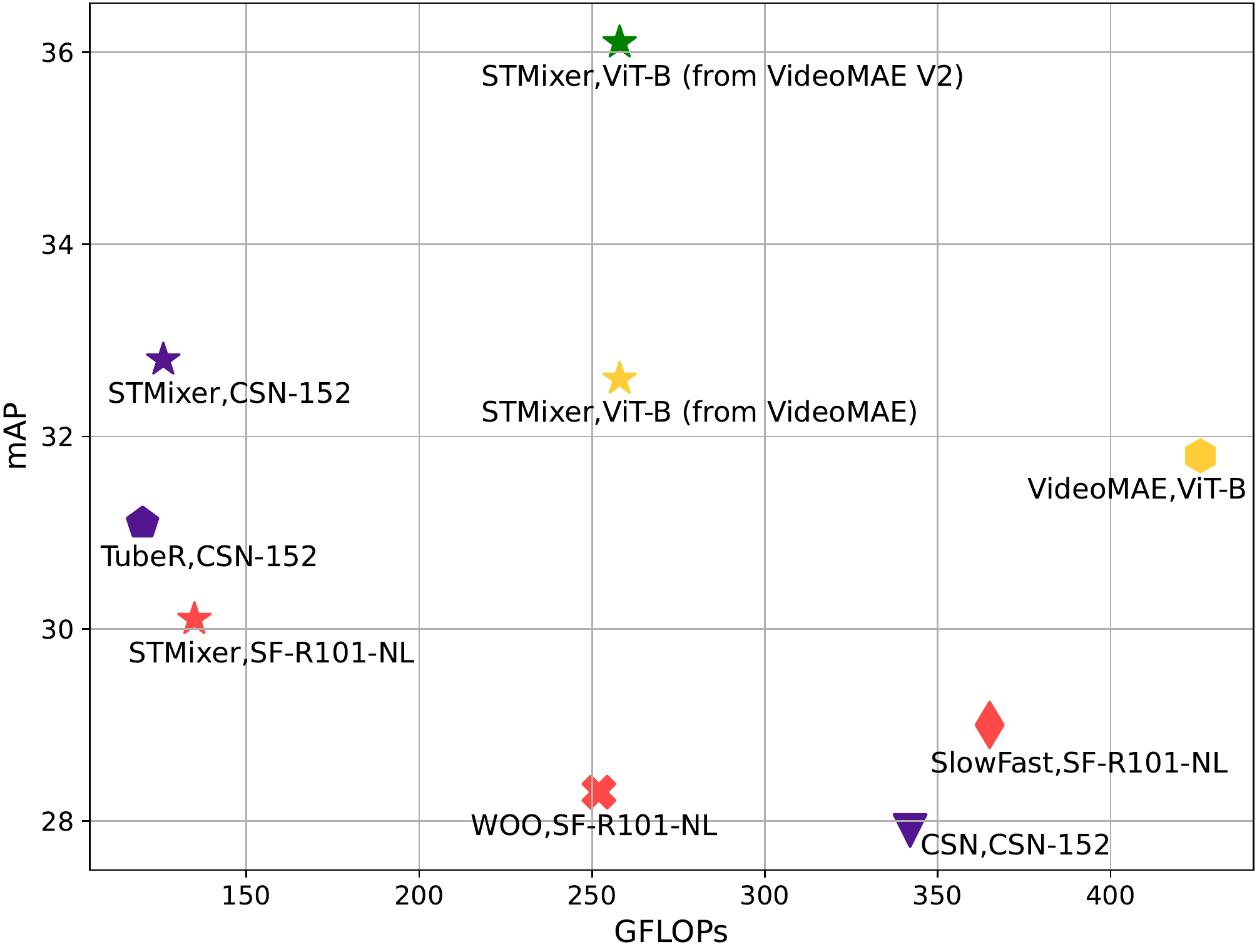}
    \vspace{-2mm}
    \caption{\textbf{Comparion of mAP versus GFLOPs.} We report detection mAP on AVA v2.2. The GFLOPs of CSN, SlowFast, and VideoMAE are the sum of Faster RCNN-R101-FPN detector GFLOPs and classifier GFLOPs. Different methods are marked by different makers and models with the same backbone are marked in the same color. The results of CSN are from \cite{TubeR}. Our STMixer achieves the best effectiveness and efficiency balance.}
    \label{fig:flops}
    \vspace{-4mm}
\end{figure}

\label{para:two_basics}
Most current action detectors adopt the two-stage Faster R-CNN-alike detection paradigm~\cite{fasterrcnn}. They share two basic designs. First, they use an auxiliary human detector to generate actor bounding boxes in advance. The training of the human detector is decoupled from the action classification network. Second, in order to predict the action category for each actor box, the RoIAlign~\cite{maskrcnn} operation is applied on video feature maps to extract actor-specific features. However, these two-stage action detection pipeline has several critical issues. First, it requires multi-stage training of person detector and action classifier, which requires large computing resources. Furthermore, the RoIAlign~\cite{maskrcnn} operation constrains the video feature sampling inside the actor bounding box and lacks the flexibility of capturing context information in its surroundings. To enhance RoI features, recent works use an extra heavy module that introduces interaction features of context or other actors~\cite{aia,acarn}.

Recently sparse query-based object detector~\cite{DETR,deformableDETR,sparsercnn} has brought a new perspective on detection tasks. Several query-based sparse action detectors~\cite{woo,TubeR} are proposed. The key idea is that action instances can be represented as a set of learnable queries, and detection can be formulated as a set prediction task, which could be trained by a matching loss. These query-based methods detect action instances in an end-to-end manner, thus saving computing resources. However, the current sparse action detectors still lack adaptability in feature sampling or feature decoding, thus suffering from inferior accuracy or slow convergence issues.

For example, building on the DETR~\cite{DETR} framework, TubeR~\cite{TubeR} adaptively attends action-specific features from single-scale feature maps but perform feature transformation in a static mannner. On the contrary, though decoding sampled features with dynamic interaction heads, WOO~\cite{woo} still uses the 3D RoIAlign~\cite{maskrcnn} operator for feature sampling, which constrains feature sampling inside the actor bounding box and fails to take advantage of other useful information in the entire spatiotemporal feature space.

Following the success of adaptive sparse object detector AdaMixer~\cite{AdaMixer} in images, we present a new query-based one-stage sparse action detector, named STMixer. 
Our goal is to create a simple action detection framework that can sample and decode features from the complete spatiotemporal video domain in a more flexible manner, while retaining the benefits of sparse action detectors, such as end-to-end training and reduced computational cost.
Specifically, we come up with two core designs. First, to overcome the aforementioned fixed feature sampling issue, we present a query-guided adaptive feature sampling module. This new sampling mechanism endows our STMixer with the flexibility of mining a set of discriminative features from the entire spatiotemporal domain and capturing context and interaction information. Second, we devise a dual-branch feature mixing module to extract discriminative representations for action detection. It is composed of an adaptive spatial mixer and an adaptive temporal mixer in parallel to focus on appearance and motion information, respectively. Coupling these two designs with a video backbone yields a simple, neat, and efficient end-to-end actor detector, which obtains a new state-of-the-art performance on the datasets of AVA~\cite{ava}, UCF101-24~\cite{ucf101}, and JHMDB~\cite{jhmdb}. In summary, our contribution is threefold:
\begin{itemize}
    \item We present a new one-stage sparse action detection framework in videos (STMixer). Our STMixer is easy to train in an end-to-end manner and efficient to deploy for action detection in a single stage.
    \item We devise two flexible designs to yield a powerful action detector. The adaptive sampling can select the discriminative feature points and the adaptive feature mixing can enhance spatiotemporal representations.
    \item STMixer achieves a new state-of-the-art performance on three challenging action detection benchmarks. 
\end{itemize}

\section{Related Work}

\noindent \bf Action detectors using an extra human detector. \rm Most current action detectors~\cite{lfb,slowfast,mvit,aia,acarn,hit,vmae} rely on an auxiliary human detector to perform actor localization on the keyframes. Typically, the powerful Faster RCNN-R101-FPN~\cite{fasterrcnn} detector is used as the human detector, which is first pre-trained on the COCO~\cite{coco} dataset and then fine-tuned on the AVA~\cite{ava} dataset. With actor bounding boxes predicted in advance, the action detection problem is reduced to a pure action classification problem. The RoIAlign~\cite{maskrcnn} operation is applied on the 3D feature maps extracted by a video backbone to generate actor-specific features. SlowFast~\cite{slowfast} and MViT~\cite{mvit} directly use the RoI features for action classification. However, RoI features only contain the information inside the bounding box but overlook context and interaction information outside the box. To remedy this inherent flaw of RoI features, AIA~\cite{aia} and ACARN~\cite{acarn} resort to using an extra heavy module that models the interaction between the actor and context or other actors. The models with an extra human detector require two-stage training and reference, which is computing resources unfriendly. Besides, they suffer from the aforementioned issue of fixed RoI feature sampling.

\noindent \bf End-to-end action detectors. \rm Methods of another research line use a single model to perform action detection. Most of them~\cite{yowo,acrn,vat,woo} still follow the two-stage pipeline but simplify the training process by jointly training the actor proposal network and action classification network in an end-to-end manner. These methods still have the issue of fixed RoI feature sampling. To remedy this, VTr~\cite{vat} attends RoI features to full feature maps while ACRN~\cite{acrn} introduces an actor-centric relation network for interaction modeling. Recently, several one-stage action detectors~\cite{moc,TubeR} are proposed. MOC~\cite{moc} is a point-based dense action detector, which uses an image backbone for frame feature extraction. It concatenates frame features along the temporal axis to form the video feature maps. Each point on the feature maps is regarded as an action instance proposal. The bounding box and action scores of each point are predicted by convolution. MOC~\cite{moc} relies more on appearance features, lacks temporal and interaction modeling and requires post-process. Building on DETR~\cite{DETR} framework, TubeR~\cite{TubeR} is a query-based action detector. TubeR~\cite{TubeR} adaptively samples features from single-scale feature maps, neglecting multi-scale information which is important for detection tasks. As DETR~\cite{DETR}, TubeR~\cite{TubeR} transforms features in a static manner, resulting in slower convergence.

Inspired by AdaMixer~\cite{AdaMixer}, we propose a new one-stage query-based detector for video action detection. Different from former query-based action detectors~\cite{woo,TubeR}, we adaptively sample discriminative features from a multi-scale spatiotemporal feature space and decode them with a more flexible scheme under the guidance of queries. 

\begin{figure*}[ht]
    \centering
    \includegraphics[width=15.5cm]{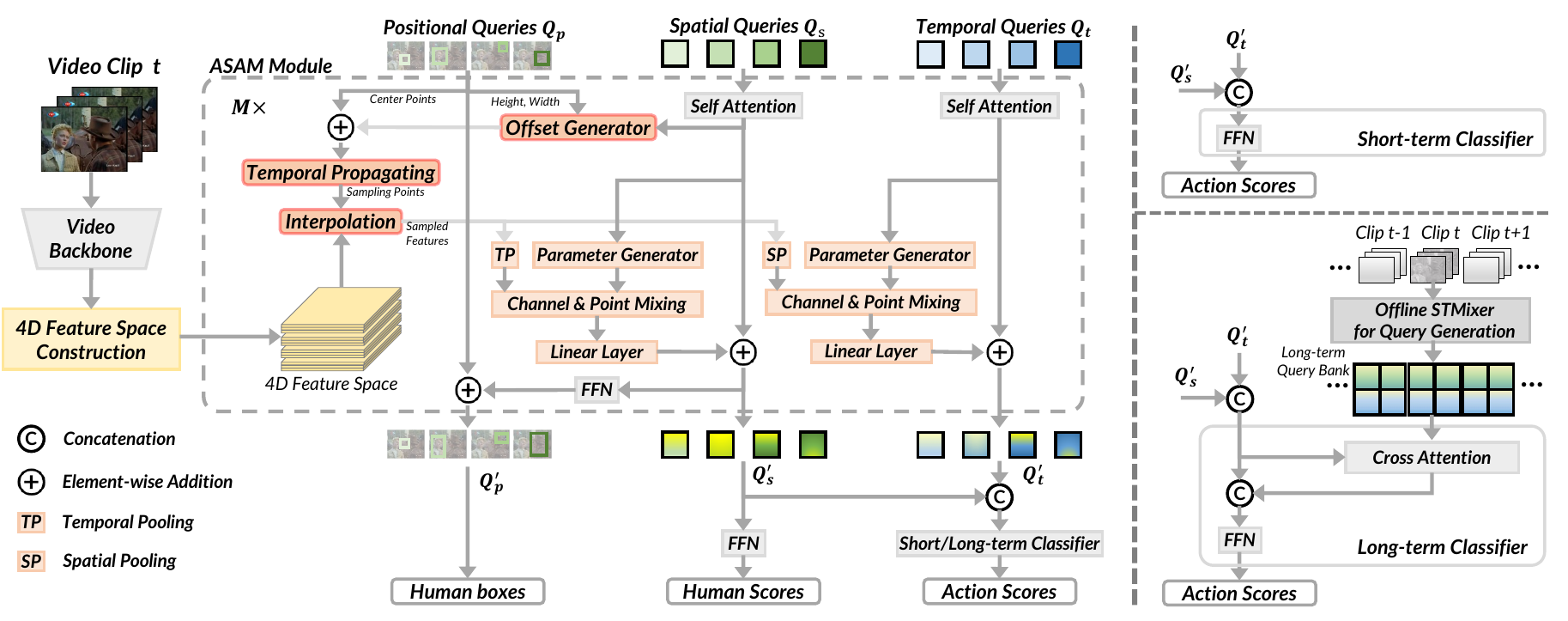}
    \scriptsize
    \vspace{-1.5mm}
    \caption{{\bf Pipeline of STMixer}. In the left, we present the overall STMixer framework. A video clip is input to the video backbone for feature extraction and a 4D feature space is constructed based on the feature maps (see Section~\ref{para:feature_space}). Then, a decoder containing $M$ ASAM modules iteratively performs adaptive feature sampling (see Section~\ref{para:adasample}) from the 4D feature space and adaptive mixing (see Section~\ref{para:adamix}) on the sampled features under the guidance of a set of learnable queries. Inversely, the queries are updated with mixed features. Optionally, a short-term or long-term classifier can be used for action scores prediction, whose detailed structures are illustrated in the right. The long-term classifier refers to the long-term query bank produced by an offline STMixer for long-term information (see Section~\ref{para:longterm}).}
    \label{fig:framework}
\end{figure*}

\section{Method}
This section presents our one-stage sparse action detector, called STMixer. The overall pipeline of STMixer is shown in Figure~\ref{fig:framework}. Our STMixer comprises a video backbone for feature extraction, a feature space construction module, and a sparse action decoder composed of $M$ adaptive sampling and adaptive mixing (ASAM) modules followed by prediction heads.  As a common practice, we set the middle frame of an input clip as the keyframe. We first use the video backbone to extract feature maps for the video and a 4D feature space is constructed based on the feature maps. Then, we develop the sparse action decoder with a set of learnable queries. Under the guidance of these queries, we perform adaptive feature sampling and feature mixing in the 4D feature space. The queries are updated iteratively. Finally, we decode each query as a detected action instance of action scores, human scores, and a human box. We will describe the technical details of each step in the next subsections.

\subsection{4D Feature Space Construction} \label{para:feature_space}

\begin{figure}[t]
    \centering
    \includegraphics[width=7cm]{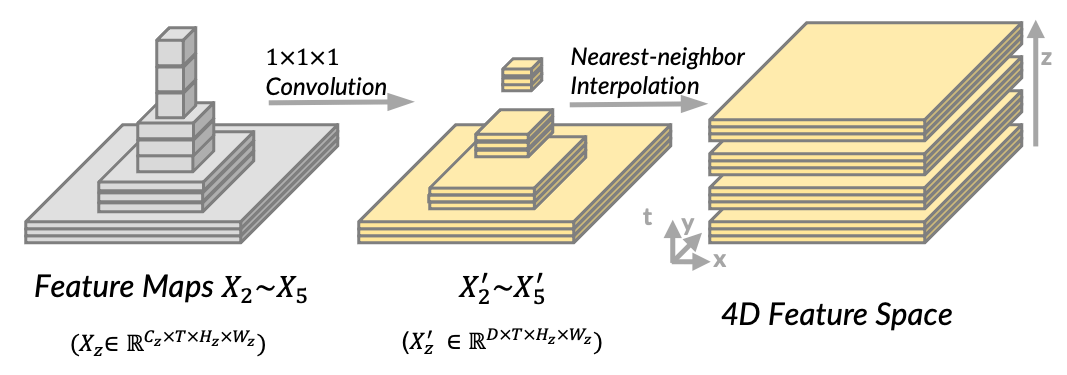}
    \vspace{-2mm}
    \caption{\textbf{4D feature space construction for hierarchical video backbone.} We construct 4D feature space on multi-scale 3D feature maps from hierarchical video backbone by simple lateral convolution and nearest-neighbor interpolation. The four dimensions of the 4D feature space are x-, y-, t-axis, and scale index z. }
    \label{fig:featurespace}
    \vspace{-3mm}
\end{figure}

\noindent \bf Hierarchical video backbone. \rm Formally, let $X_z \in \mathbb{R}^{C_z\times T\times H_z\times W_z}$ denote the feature map of convolution stage $z$ of the hierarchical backbone, where $z \in \{2,3,4,5\}$, $C_z$ stands for the channel number, $T$ for time, $H_z$ and $W_z$ for the spatial height and width. The stage index $z$ can be seen as the scale index of the feature map as $X_z$ has the downsampling rate of $2^z$. We first transform each feature map $X_z$ to the same channel $D$ by $1 \times 1\times 1$ convolution. Then, we rescale the spatial shape of each stage feature map to $H_2 \times W_2$ by simple nearest-neighbor interpolation and align them along the x- and y-axis. The four dimensions of the constructed feature space are x-, y-, t-axis, and scale index z, respectively. This process is illustrated in Figure~\ref{fig:featurespace}.

\noindent \bf Plain ViT backbone. \rm To make STMixer framework compatible with plain ViT~\cite{ViT} backbone, inspired by ViTDet~\cite{VitDet}, we construct 4D feature space based on the last feature map from ViT backbone. Specifically, with the output feature map of the default downsampling rate of $2^4$, we first produce hierarchical feature maps $\left\{X_{z}\right\}$ of the same channel number $D$ using convolutions of spatial strides $\left\{\frac{1}{4},\frac{1}{2},1,2 \right\}$, where a fractional stride indicates a deconvolution. Then we also rescale each feature map to the spatial size of $H_2 \times W_2$.

\subsection{Query Definition} \label{para:query_def}

The definition of our queries derives from the object query in Sparse R-CNN~\cite{sparsercnn}, but we specify the action query in a disentangled fashion. Specifically, we factorize action queries into spatial queries $Q_s \in \mathbb{R}^{N\times D}$ and temporal queries $Q_t \in \mathbb{R}^{N\times D}$ to refer to the spatial content and temporal content respectively. $N$ represents the number of queries, while $D$ denotes the dimension of each query.

We further define positional queries $Q_p \in \mathbb{R}^{N\times 4}$. Each positional query $Q{}_{p}^{n}$ ($n$ stands for the query index) is formulated as a proposal box vector $(x^n,y^n,z^n,r^n)$. Formally, $x^n$ and $y^n$ stand for the x- and y-axis coordinates of the box center, and $z^n$ and $r^n$ denote the logarithm of its scale (\ie the area of the bounding box) and aspect ratio. The positional queries $Q_p$ are initialized in such a way that every box vector is the whole keyframe. 

\subsection{Adaptive Spatiotemporal Feature Sampling} \label{para:adasample}

Different from previous work~\cite{slowfast,aia,woo,acarn} that samples RoI features by pre-computed proposal boxes, we sample actor-specific features adaptively from the aforementioned 4D feature space under the guidance of spatial queries. Specifically, given a spatial query $Q{}_{s}^{n}$ and the corresponding positional query $Q{}_{p}^{n}$, we regard the center point $(x^n,y^n,z^n)$ of the proposal box as the reference point, and regress $P_{in}$ groups of offsets along x-, y-axis and scale index z on query $Q{}^{n}_{s}$, using a linear layer:
\begin{equation}
\begin{gathered}
    \left\{ \left( \bigtriangleup x^{n}_{i},\bigtriangleup y^{n}_{i},\bigtriangleup z^{n}_{i} \right)  \right\} = \text{Linear} (Q{}^{n}_{s}),\\
    \text{where}\ i\in \mathbb{Z}\ \text{and}\ 1\leqslant i\leqslant P_{in},
\end{gathered}
\end{equation}
where $P_{in}$ is the number of sampled points for each query. Then, the offsets are added to the reference point, thus $P_{in}$ spatial feature points are obtained:
\begin{equation}
    \begin{cases}\widetilde{x}^{n}_{i}=x^n+\bigtriangleup x^{n}_{i}\cdot 2^{z^n-r^n},\\ 
    \widetilde{y}^{n}_{i}=y^n+\bigtriangleup y^{n}_{i}\cdot 2^{z^n+r^n},\\ 
    \widetilde{z}^{n}_{i}=z^n+\bigtriangleup z^{n}_{i}.\end{cases}  
\end{equation}
where $2^{z^n-r^n}$ and $2^{z^n+r^n}$ are the width and height of the box respectively. We offset the spatial position of sampling points with respect to the width and height of the box to reduce the learning difficulty.

Finally, we propagate sampling points along the temporal axis, thus obtaining $T\times P_{in}$ points to sample from the 4D feature space. In our implementation,  we simply copy these spatial sampling points along the temporal dimension because current action detection datasets yield {\em temporal slowness property} that the variation of actor locations along the temporal dimension is very slow. We compare different ways of temporal propagating in our ablation study.

Given $T\times P_{in}$ sampling points, we sample instance-specific features by interpolation from the 4D feature space. In the following sections, the sampled spatiotemporal feature for spatial query $Q_s^n$ is denoted by $F^{n} \in \mathbb{R}^{T\times P_{in}\times D}$. 

\subsection{Adaptive Dual-branch Feature Mixing} \label{para:adamix}

After feature sampling, we factorize the sampled features into spatial features and temporal features by pooling and then enhance each by adaptive mixing respectively. As the dual-branch mixing module is completely symmetrical, we only describe spatial mixing in detail and temporal mixing is performed in a homogeneous way.

Different from MLP-Mixer~\cite{mlpmixer}, our mixing parameters are generated adaptively. Given a spatial query $Q_s^n \in \mathbb{R}^{D}$ and its corresponding sampled features $F^n \in \mathbb{R}^{T\times P_{in}\times D}$, we first use a linear layer to generate query-specific channel-mixing weights $M_c \in \mathbb{R}^{D\times D}$, and then apply plain matrix multiplication on temporally pooled feature and $M_c$ to perform channel-mixing, given by:
\begin{equation}
    M_c = \text{Linear}(Q_s^n) \in \mathbb{R}^{D\times D},
\end{equation}
\begin{equation}
    \text{CM}(F^n) = \text{ReLU}(\text{LayerNorm}(\text{GAP}(F^n) \times M_c)).
\end{equation}
where GAP stands for the global average pooling operation in the temporal dimension while LayerNorm for the layer normalization~\cite{layernorm}. We use CM$(F^n)\in \mathbb{R}^{P_{in} \times D}$ to denote the channel-wise mixing output features.

After channel-mixing, we perform point-wise mixing in a similar way. Suppose $P_{out}$ is the number of spatial point-wise mixing out patterns, we use PCM$(F^n)\in \mathbb{R}^{D\times P_{out}}$ to denote the point-wise mixing output features, given by:
\begin{equation}
    M_p = \text{Linear}(Q_s^n) \in \mathbb{R}^{P_{in}\times P_{out}},
\end{equation}
\begin{equation}
    \text{PCM}(F^n) = \text{ReLU}(\text{LayerNorm}({\text{CM}(F^n)}^T \times M_p)).
\end{equation}

The final output PCM$(F^n)$ is flattened, transformed to $D$ dimension, and added to the spatial query $Q^n_s$. The temporal query $Q^n_t$ is updated in a homogeneous way, except that the pooling is applied in the spatial dimension. After global spatial pooling, there are $T$ feature points of different temporal locations for temporal mixing and we set the number of temporal point-wise mixing out patterns to $T_{out}$.

\subsection{Sparse Action Decoder}
STMixer adopts a unified action decoder for both actor localization and action classification. The decoder comprises $M$ stacked ASAM modules followed by a feed-forward network (FFN) for human scores prediction and a short-term or long-term classifier for action score prediction. In this section, we represent the structure of ASAM module and specify the outputs of the prediction heads.

\noindent \bf ASAM module. \rm \label{para:ASAM} The overall structure of ASAM module is shown in Figure \ref{fig:framework}. We first perform self-attention~\cite{attention} on spatial queries $Q_s$ and temporal queries $Q_t$ to capture the relation information between different instances. Then we perform adaptive sampling from the 4D feature space and dual-branch adaptive mixing on the sampled feature as described before. The spatial queries $Q_s$ and temporal queries $Q_t$ are updated with mixed features. An FFN is applied on updated spatial queries $Q_s^{\prime}$ to update the positional queries $Q_p$. The updated queries $Q_p^{\prime}$, $Q_s^{\prime}$ and $Q_t^{\prime}$ are used as inputs for the next ASAM module.

\noindent \bf Outputs of the prediction heads. \rm The output human boxes are decoded from positional queries $Q_p^{\prime}$. We apply an FFN on $Q_s^{\prime}$ to predict human scores $S_{H} \in \mathbb{R}^{N\times 2}$ which indicates the confidence values that each box belongs to the human category and background. Based on the concatenation of spatial queries $Q_s^{\prime}$ and temporal queries $Q_t^{\prime}$, we use a short-term or long-term classifier to predict action scores $S_{A} \in \mathbb{R}^{N\times C}$, where $C$ is the number of action classes. 

\noindent \bf Short-term and long-term classifier. \rm \label{para:longterm} Short-term classifier (see Figure~\ref{fig:framework} right top) is a simple FFN which predicts action scores based on short-term information of current queries while long-term classifier (see Figure~\ref{fig:framework} right bottom) refers to long-term query bank for long-term information. Our design of the long-term classifier is adapted from LFB~\cite{lfb}. We train an STMixer model without long-term information first. Then, for a video of $\mathcal{T}$ clips, we use the trained STMixer to perform inference on each clip of it. We store the concatenation of the spatial and temporal queries from the last ASAM module corresponding to the $k$ highest human scores in the query bank for each clip. We denote the stored queries for clip of time-step $t$ as $L_t\in \mathbb{R}^{k\times d}$ where $d=2D$, and the long-term query bank of the video as $L=[L_0,L_1,...,L_{\mathcal{T}-1}]$. Given the long-term query bank of all videos, we train an STMixer model with a long-term classifier from scratch. For video clip $t$, we first sample a window of length $w$ from the long-term query bank centered at it and stack this window into $\tilde{L_{t}}\in \mathbb{R}^{K\times d}$:
\begin{equation}
    \tilde{L_{t}} = \text{stack}([L_{t-w/2},..., L_{t+w/2-1} ]).
\end{equation}
We then infer current queries to $\tilde{L_{t}}$ for long-term information by cross-attention~\cite{attention}:
\begin{equation}
    S_t^\prime = \text{cross-attention}(S_t,\tilde{L_{t}}),
\end{equation}
where $S_t\in \mathbb{R}^{N\times d}$ is the concatenation of the output spatial queries and temporal queries of the last ASAM module. The output $S_t^\prime$ is then channel-wise concatenated with $S_t$ for action scores prediction.

\subsection{Training}

We compute training loss based on the output human boxes, human scores, and action scores. Consistent with WOO~\cite{woo}, the loss function $\mathcal{L}$ consists of set prediction loss $\mathcal{L}_{set}$~\cite{DETR,deformableDETR,sparsercnn} and action classification loss $\mathcal{L}_{act}$. Formally, 
\begin{equation}
    \mathcal{L}_{set}=\lambda_{cls}\mathcal{L}_{cls}+\lambda_{L_1}\mathcal{L}_{L_1}+\lambda_{giou}\mathcal{L}_{giou},
\end{equation}
\begin{equation}
    \mathcal{L}=\mathcal{L}_{set}+\lambda_{act}\mathcal{L}_{act}.
\end{equation}
 $\mathcal{L}_{cls}$ denotes the cross-entropy loss over two classes (human and background). $\mathcal{L}_{L_1}$ and $ \mathcal{L}_{giou}$ are box loss inherited from \cite{DETR,deformableDETR,sparsercnn}. As in \cite{DETR,woo}, we first use Hungarian algorithm to find an optimal bipartite matching between predicted actors and ground truth actors according to $\mathcal{L}_{set}$. Then we calculate full training loss $\mathcal{L}$ based on the matching results. $\mathcal{L}_{act}$ is binary cross entropy loss for action classification. We only compute $\mathcal{L}_{act}$ for the prediction matched with a ground truth. $\lambda_{cls}$, $\lambda_{L_1}$, $\lambda_{giou}$, and $\lambda_{act}$ are corresponding weights of each term.

\section{Experiments}

\subsection{Experimental Setup}
\noindent \bf Datasets. \rm The AVA dataset~\cite{ava} contains 211k video clips for training and 57k for validation segmented from 430 15-minute videos. The videos are annotated at 1FPS over 80 atomic action classes. Following the standard evaluation protocol~\cite{ava}, we report our results on 60 action classes that have at least 25 validation examples.
JHMDB~\cite{jhmdb} consists of 928 temporally trimmed videos from 21 action classes. Results averaged over three splits are reported. UCF101-24 is a subset of UCF101~\cite{ucf101}. It contains 3,207 videos annotated with action instances of 24 classes. As the common setting, we report the performance on the first split.

\noindent \bf Network configurations. \rm We configure the dimension $D$ of both spatial and temporal queries to 256 and set the number of both queries $N$ equaling 100. The number of sampling point $P_{in}$ in each temporal frame is set to 32. The spatial and temporal mixing out patterns $P_{out}$ and $T_{out}$ are set to 4 times the number of sampling points and temporal frames respectively, that is, 128 and 32 for SlowFast~\cite{slowfast} backbone and 128 and 16 for CSN~\cite{csn} and ViT~\cite{ViT} backbone. Following multi-head attention~\cite{attention} and group convolution~\cite{groupconv}, we split the channel $D$ into 4 groups and perform group-wise sampling and mixing. We stack 6 ASAM modules in our action decoder as default. For the long-term classifier, we set the number of stored queries of each clip $k$ as 5 and window length $w$ as 60. The number of cross-attention layers is set to 3.

\noindent \bf Losses and optimizers. \rm We set the loss weight in STMixer as $\lambda_{cls} = 2.0$, $\lambda_{L_1} = 2.0$, $\lambda_{giou} = 2.0$ and $\lambda_{act} = 24.0$. We use AdamW~\cite{adamw} optimizer with weight decay $1 \times 10^{-4}$ for all experiments. Following \cite{DETR,woo}, intermediate supervision is applied after each ASAM module. 

\noindent \bf Training and inference recipes. \rm We train STMixer detectors for 10 epochs with an initial learning rate of $2.0 \times 10^{-5}$ and batchsize of 16. The learning rate and batchsize can be tuned according to the linear scaling rule~\cite{linearscale}. We randomly scale the short size of the training video clips to 256 or 320 pixels. Color jittering and random horizontal flipping are also adopted for data augmentation.

For inference, we scale the short size of input frames to 256 as the common setting. Given an input video clip, STMixer predicts $N$ bounding boxes associated with their human scores and action scores from the last ASAM module. If the confidence score that a box belongs to the human category is higher than the preset threshold, we take it as a detection result. We set the threshold to 0.6 for AVA. The performances are evaluated with official metric frame-level mean average precision(mAP) at 0.5 IoU threshold.

\subsection{Ablation Study} 
\label{subsection:ablation}
\begin{table*}[h]
\footnotesize
\begin{subtable}[t]{0.33\linewidth}
\centering
\vspace{-17.5pt}
\resizebox{\linewidth}{!}{
\begin{tabular}{cc c c}
\Xhline{1.2pt}
\textbf{Classification}       & \textbf{Localization}   & mAP & GFLOPs \\ \Xhline{1.2pt}
\rowcolor[HTML]{EFEFEF} 
\multicolumn{2}{c}{4D Feature Space} & 23.1 & 44.4       \\
Key-Frame Features   & Res5 Features  & 22.8 & 53.7       \\ \Xhline{1.2pt}
\end{tabular}

}
\caption{\textbf{Feature space.} Sampling features from 4D feature space is more effective and efficient than sampling from key-frame features for classification and res5 features for localization.}
\label{tab:featspace}
\end{subtable}
\hfill
\begin{subtable}[t]{0.25\linewidth}
\centering
\vspace{-17.5pt}
\resizebox{0.98\linewidth}{!}{
\begin{tabular}{l c c}
\Xhline{1.2pt}
                           & mAP & GFLOPs \\ \Xhline{1.2pt}
\rowcolor[HTML]{EFEFEF} 
Simple Lateral Conv. & 23.1 & 44.4  \\
Full FPN                   & 22.9 & 45.8  \\ \Xhline{1.2pt}
\end{tabular}
}
\caption{\textbf{4D feature space construction.} Simple lateral convolution achieves comparable performance while being more efficient than using full FPN.}
\label{tab:featspacecons}
\end{subtable}
\hfill
\begin{subtable}[t]{0.38\linewidth}
\centering
\resizebox{\linewidth}{!}{
\begin{tabular}{l c c c}
\Xhline{1.2pt}
\textbf{Sampling Strategy}                    & $P_{in}$  & mAP & GFLOPs \\ \Xhline{1.2pt}
Fixed Grid Sampling                  & 49 & 21.9 & 
45.6 \\
\rowcolor[HTML]{EFEFEF} 
Adaptive Sampling + Temporal Copying & 32 & 23.1 & 44.4 \\
Adaptive Sampling + Temporal Moving  & 32 & 23.3 &   44.6 \\ \Xhline{1.2pt}
\end{tabular}
}
\caption{\textbf{Sampling strategy.} Sampling fewer feature points, our adaptive sampling strategy achieves better performance than fixed grid sampling. Temporal moving brings slight improvement in mAP.}
\label{tab:samplestrategy}
\end{subtable}

\begin{subtable}[t]{0.33\linewidth}
\centering
\vspace{-4pt}
\resizebox{0.89\linewidth}{!}{
\begin{tabular}{c c c >{\columncolor[HTML]{EFEFEF}}c c c}
\Xhline{1.2pt}
$P_{in}$ & 8 & 16 & 32 & 48 & 64 \\ \Xhline{1.2pt}
mAP      & 22.4 & 22.5 & 23.1 & 23.0 & 22.5 \\ 
GFLOPs   & 42.4 & 43.1 & 44.4 & 45.6 & 46.9 \\ \Xhline{1.2pt}
\end{tabular}
}
\caption{\textbf{Number of sampling points.} Sampling 32 points per frame achieves the best performance.}
\label{tab:samplenum}
\end{subtable}
\hfill
\begin{subtable}[t]{0.25\linewidth}
\centering
\vspace{-4pt}
\resizebox{\linewidth}{!}{
\begin{tabular}{c c c >{\columncolor[HTML]{EFEFEF}}c c}
\Xhline{1.2pt}
$N$     & 15    & 50    & 100    & 150    \\ \Xhline{1.2pt}
mAP     & 22.1  & 22.9  & 23.1   & 22.5   \\
GFLOPs  & 31.7  & 36.9  & 44.4   & 51.8   \\ \Xhline{1.2pt}
\end{tabular}
}
\caption{\textbf{Number of queries.} Using 100 queries works the best.}
\label{tab:numquery}
\end{subtable}
\hfill
\begin{subtable}[t]{0.38\linewidth}
\centering
\vspace{5pt}
\resizebox{0.9\linewidth}{!}{

\begin{tabular}{l c c}
\Xhline{1.2pt}
\textbf{Mixing Strategy }                  & mAP  & GFLOPs \\ \Xhline{1.2pt}
fixed parameter dual-branch mixing            & 22.5 & 36.8   \\
\rowcolor[HTML]{EFEFEF} 
dual-branch spatiotemporal mixing & 23.1 & 44.4   \\
spatial mixing only               & 22.4 & 40.4   \\
temporal mixing only              & 22.1 & 33.5   \\
sequential spatiotemporal mixing  & 22.6 & 43.6   \\
coupled spatiotemperal  mixing    & 22.8 & 93.2   \\
  \Xhline{1.2pt}
\end{tabular}
}
\caption{\textbf{Mixing strategy.} Query-guided adaptive feature mixing outperforms fixed parameter mixing and our dual-branch spatiotemporal feature mixing strategy works the best. }
\label{tab:mixingstrategy}
\end{subtable}

\begin{subtable}[t]{0.33\linewidth}
\centering
\vspace{-48pt}
\resizebox{1.0\linewidth}{!}{

\begin{tabular}{c c c >{\columncolor[HTML]{EFEFEF}}c c c}
\Xhline{1.2pt}
$P_{out}$/$T_{out}$  & 64/8   & 96/12   & 128/16  & 160/20  & 192/24  \\\Xhline{1.2pt}

mAP        & 22.5 & 22.8 & 23.1 & 22.9 & 22.6 \\
GFLOPs     & 40.2 & 42.3 & 44.4 & 46.4 & 48.4 \\ \Xhline{1.2pt}
\end{tabular}
}
\caption{\textbf{Number of mixing out patterns.} A moderate number of mixing out patterns works the best. }
\label{tab:outpatterns}
\end{subtable}
\quad
\begin{subtable}[t]{0.25\linewidth}
\centering
\vspace{-48.5pt}
\resizebox{0.95\linewidth}{!}{
\begin{tabular}{c c c >{\columncolor[HTML]{EFEFEF}}c c}
\Xhline{1.2pt}
$M$     & 1    & 3    & 6    & 9    \\ \Xhline{1.2pt}
mAP     & 18.4 & 22.5 & 23.1 & 22.6 \\
GFLOPs  & 32.0 & 36.9 & 44.4 & 51.8 \\ \Xhline{1.2pt}
\end{tabular}
}
\caption{\textbf{Number of ASAM modules.} Using 6 ASAM modules works the best.}
\label{tab:nummodules}
\end{subtable}

\vspace{-3pt}
\caption{\textbf{Ablations Experiments.} We use a SlowOnly ResNet-50 backbone to perform our ablation studies. Models are trained on the training set of AVA v2.2 and evaluated on the validation set. Default choices for our model are colored in \colorbox[HTML]{EFEFEF}{gray}. }
\label{tab:ablations}
\vspace{-2mm}
\end{table*}

We conduct ablation experiments on AVA v2.2 dataset to investigate the influence of different components in our STMixer framework. A SlowOnly ResNet-50 backbone~\cite{fan2020pyslowfast} is used for our ablation experiments. We report both mAP and GFLOPs for effectiveness and efficiency comparison.

\noindent \bf Ablations on 4D feature space. \rm We first show the benefit of sampling features from the unified 4D feature space. For comparison, we design a two-stage counterpart of our STMixer. In the first stage, we sample features from key-frame features for actor localization. In the second stage, we sample features from res5 features for action classification. As shown in Tabel~\ref{tab:featspace}, sampling features from a unified 4D feature space and performing action detection in a one-stage manner significantly reduce computing costs, and detection mAP also gets improved as multi-scale information is also utilized for classification. We then investigate two different ways for the 4D feature space construction. As shown in Table~\ref{tab:featspacecons}, constructing 4D feature space by simple lateral $1\times 1\times 1$ convolution achieves comparable detection accuracy while being more efficient than using full FPN~\cite{fpn} with a top-down pathway.  

\noindent \bf Ablations on feature sampling. \rm In Tabel~\ref{tab:samplestrategy}, we compare 3 different feature sampling strategies. For fixed grid feature sampling, we sample $7\times7$ feature points inside each actor box by interpolation, which is actually the RoIAlign~\cite{maskrcnn} operation adopted by many former methods~\cite{woo,vat}. Though sampling fewer points per group, our adaptive sampling strategy improves the detection mAP by 1.2. The results show that the fixed RoIAlign operation for feature sampling fails to capture useful context information outside the box while our adaptive sampling strategy enables the detector to mine discriminative features from the whole 4D feature space. Beyond simply copying sampling points, we try sampling different feature points in different frames by predicting the offset of the reference bounding box for each frame. The improvement is marginal due to the slowness issue of current action detection datasets that the variation of actors' location and action is very slow. In Table~\ref{tab:samplenum}, we further investigate the influence of the number of sampling points per group 
 $P_{in}$ in each temporal frame. Setting $P_{in}=32$ achieves the best detection performance.

\noindent \bf Ablations on feature mixing. \rm In Table~\ref{tab:mixingstrategy}, we compare different feature mixing strategies. We first demonstrate the benefit of query-guided adaptive mixing. The detection mAP drops by 0.6 when using fixed parameter mixing. For adaptive mixing, the mixing parameters are dynamically generated based on a specific query, thus more powerful to enhance the presentation of each specific action instance proposal. We further compare different adaptive feature mixing strategies. From the results in Table~\ref{tab:mixingstrategy}, it is demonstrated that both spatial appearance and temporal motion information are important for action detection. However, coupled spatiotemporal feature mixing using queries of a single type has a high computational complexity. Decoupling features along spatial and temporal dimensions saves computing costs. Our dual-branch spatiotemporal feature mixing outperforms sequential spatiotemporal feature mixing by 0.5 mAP. This is because actor localization at keyframes only needs spatial appearance information and parallel dual-branch mixing will reduce the effect of temporal information on localization. Also, by concatenating temporal queries to spatial queries, more temporal information is leveraged for action classification. In Table~\ref{tab:outpatterns}, we investigate different spatial and temporal mixing out patterns $P_{out}$ and $T_{out}$ from 64 and 8 to 192 and 24, that is, 2 times to 6 times the number of sampling points and temporal frames. Setting $P_{out}$ and $T_{out}$ equaling 128 and 16 achieves the best performance. 

\noindent \bf Ablations on network configuration. \rm As shown in Table~\ref{tab:numquery}, the detection mAP is saturated when the number of queries is increased to 100. From the results in Table~\ref{tab:nummodules}, a stack of 6 ASAM modules achieves a good effectiveness and efficiency balance.

\subsection{Comparison with State-of-the-arts on AVA}

\begin{table*}[t]
\tiny
\centering
\resizebox{0.85\linewidth}{!}{
\begin{tabular}{lcccccccc}
\Xhline{0.7pt}
\multicolumn{1}{c}{}  &         &             &              &            &        & & \multicolumn{2}{c}{\textbf{mAP}} \\ \cline{8-9} 
\multicolumn{1}{c}{\multirow{-2}{*}{\textbf{Method}}} &
  \multirow{-2}{*}{\textbf{Detector}} &
  \multirow{-2}{*}{\textbf{One-stage}} &
  \multirow{-2}{*}{\textbf{Input}} &
  \multirow{-2}{*}{\textbf{Backbone}} &
  \multirow{-2}{*}{\textbf{Pre-train}} &
  \multirow{-2}{*}{\textbf{LF}} &
  \textbf{v2.1} &
  \textbf{v2.2} \\ \Xhline{0.7pt}
\rowcolor[HTML]{EFEFEF} 
\multicolumn{8}{l}{\cellcolor[HTML]{EFEFEF}\textbf{Compare to methods with an extra human detector}} &
  \multicolumn{1}{l}{\cellcolor[HTML]{EFEFEF}} \\
SlowFast~\cite{slowfast}              & \cmark & \xmark  & 32$\times$2  & SF-R101-NL   & K600       & \xmark & 28.2            & 29.0             \\
LFB~\cite{lfb}                   & \cmark & \xmark & 32$\times$2  & I3D-R101-NL  & K400       & \cmark & 27.7            & -              \\
CA-RCNN~\cite{carcnn}               & \cmark  & \xmark& 32$\times$2  & R50-NL       & K400       & \cmark & 28.0            & -              \\
AIA~\cite{aia}                   & \cmark  & \xmark& 32$\times$2  & SF-R101      & K700       & \cmark & 31.2            & 32.3           \\
ACARN~\cite{acarn}                 & \cmark & \xmark & 32$\times$2  & SF-R101      & K400       & \cmark & 30.0            & -              \\
ACARN~\cite{acarn}                 & \cmark & \xmark & 32$\times$2  & SF-R101-NL   & K600       & \cmark & -               & 31.4           \\
VideoMAE~\cite{vmae}              & \cmark  & \xmark& 16$\times$4  & ViT-B        & K400       & \xmark & -               & 31.8           \\
VideoMAE~\cite{vmae}              & \cmark  & \xmark& 16$\times$4  & ViT-L        & K700       & \xmark & -               & 39.3           \\ \hline
STMixer                  & \xmark  & \cmark& 32$\times$2  & SF-R101      & K700       & \xmark & 30.6            & 30.9           \\
STMixer        & \xmark  & \cmark& 32$\times$2  & SF-R101      & K700       & \cmark & 32.6   & 32.9  \\
STMixer                  & \xmark & \cmark & 16$\times$4  & ViT-B (from~\cite{vmae})        & K400       & \xmark & -               & 32.6           \\
STMixer                  & \xmark & \cmark & 16$\times$4  & ViT-B (from~\cite{VideoMAEv2})      & K710+K400       & \xmark & -               & 36.1           \\
STMixer        & \xmark  & \cmark & 16$\times$4  & ViT-L  (from~\cite{vmae})      & K700  & \xmark & -               & \textbf{39.5}  \\
\Xhline{0.7pt} 
\rowcolor[HTML]{EFEFEF} 
\multicolumn{8}{l}{\cellcolor[HTML]{EFEFEF}\textbf{Compare to end-to-end methods}} &                \\
AVA~\cite{ava}                   & \xmark  & \xmark& 20$\times$1  & I3D-VGG      & K400       & \xmark & 14.6            & -              \\
ACRN~\cite{acrn}                 & \xmark  & \xmark& 20$\times$1  & S3D-G        & K400       & \xmark & 17.4            & -              \\
STEP~\cite{step}                  & \xmark  & \xmark& 12$\times$1  & I3D-VGG      & K400       & \xmark & 18.6            & -              \\
VTr~\cite{vat}                   & \xmark  & \xmark& 64$\times$1  & I3D-VGG      & K400       & \xmark & 24.9            & -              \\
WOO~\cite{woo}                   & \xmark  & \xmark& 32$\times$2  & SF-R50       & K400       & \xmark & 25.2            & 25.4           \\
WOO~\cite{woo}                   & \xmark  & \xmark& 32$\times$2  & SF-R101-NL   & K600       & \xmark & 28.0            & 28.3           \\
TubeR~\cite{TubeR}                 & \xmark  & \cmark& 32$\times$2  & CSN-152      & IG-65M     & \xmark & 29.7            & 31.1           \\
TubeR~\cite{TubeR}                 & \xmark  & \cmark& 32$\times$2  & CSN-152      & IG-65M     & \cmark & 31.7            & 33.4           \\ \hline
STMixer                  & \xmark & \cmark & 32$\times$2  & SF-R50       & K400       & \xmark & 27.2            & 27.8           \\
STMixer         & \xmark  & \cmark& 32$\times$2  & SF-R101-NL   & K600       & \xmark & 29.8   & 30.1  \\
STMixer                  & \xmark  & \cmark& 32$\times$2  & CSN-152      & IG-65M     & \xmark & 31.7            & 32.8           \\
STMixer        & \xmark  & \cmark& 32$\times$2  & CSN-152      & IG-65M     & \cmark & \textbf{34.4}   & 34.8  \\ \Xhline{0.7pt}
\end{tabular}
}
 \caption{\textbf{Comparisons with state-of-the-arts on validation sets of AVA v2.1 and v2.2.} \cmark of column ``Detector" denotes an extra human detector Faster RCNN-R101-FPN~\cite{fasterrcnn} is used. \cmark of column ``LF" denotes long-term features are used.} 
\label{tab:ava}
\vspace{-6pt}
\end{table*}

We compare our proposed STMixer with state-of-the-art methods on AVA v2.1 and v2.2 in Table~\ref{tab:ava}. We first compare our STMixer to methods using an extra offline human detector. Our STMixer with SlowFast-R101 backbone achieves 30.6 and 30.9 mAP when not using long-term features. With long-term query support, our STMixer reaches 32.6 and 32.9 mAP on AVA v2.1 and v2.2 respectively. To demonstrate the generalization ability of our method, we conduct experiments with ViT~\cite{ViT} backbone. Compared with the two-stage counterparts, STMixer brings performance improvements while getting rid of the dependence on an extra detector. Although ViT is considered to have a global receptive field, our adaptive sampling and decoding mechanism could serve as a supplement to improve the flexibility of the model. Compared to previous end-to-end methods, our STMixer achieves the best results. Our STMixer outperforms WOO~\cite{woo} by 2.0 and 1.7 mAP even though WOO test models at 320 resolution. STMixer also consistently outperforms TubeR~\cite{TubeR} on AVA v2.1 or v2.2, using or not using long-term features. 

We compare mAP versus GFLOPs on AVA v2.2 in Figure~\ref{fig:flops} to show the efficiency of our STMixer. AIA~\cite{aia} and ACARN~\cite{acarn} do not report their GFLOPs. As they are built on SlowFast~\cite{slowfast} framework and also use an offline human detector but introduce extra modules to model interaction, SlowFast can serve as a lower bound of complexity for them. For a fair comparison, we report results for no long-term feature version of each method. As shown in Figure~\ref{fig:flops}, due to an extra human detector being needed, SlowFast~\cite{slowfast}, CSN~\cite{csn}, and VideoMAE~\cite{vmae} have much higher GFLOPs than end-to-end methods with same backbone. Among end-to-end methods, STMixer achieves the best effectiveness and efficiency balance. STMixer outperforms WOO~\cite{woo} by 1.8 mAP while having much lower GFLOPs (135 versus 252). With a slight GFLOPs increase (126 versus 120), STMixer outperforms TubeR~\cite{TubeR} by 1.7 mAP.

\begin{figure*}[t]
    \centering
    \includegraphics[width=17cm]{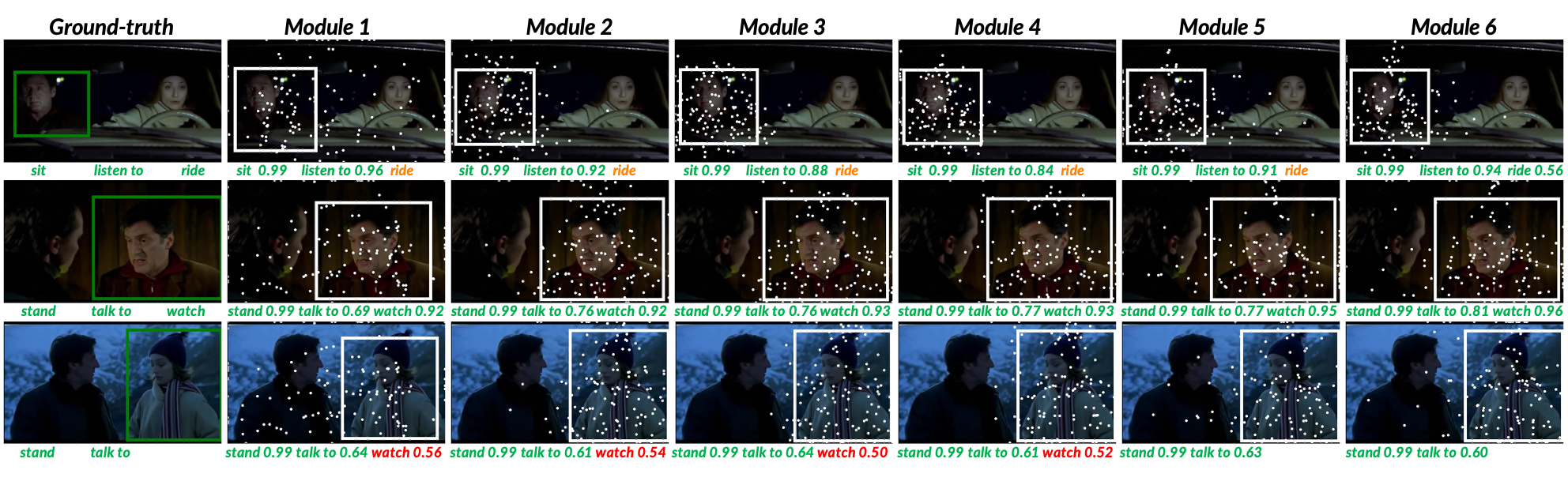}
    \vspace{-3mm}
    \caption{\textbf{Sampling points and detection results visualization.} We display the actor bounding boxes and action classes of three ground-truth action instances in the first column. Each ASAM module's sampling points, predicted actor bounding boxes and action scores are displayed in the following columns. The correctly predicted action classes are displayed in \textcolor[HTML]{23B051}{green}, missing in \textcolor{orange}{orange}, and wrongly predicted in \textcolor{red}{red}. Intuitively, STMixer mines discriminative context features outside the bounding box for better action detection.}
    \label{fig:vis}
    \vspace{-13pt}
\end{figure*}

\subsection{Results on JHMDB and UCF101-24}

\begin{table}[t]
    \centering
    \LARGE
    \resizebox{0.95\linewidth}{!}{
    \begin{tabular}{l c c c c c}
        \Xhline{1.5pt}
        \textbf{Method} &\textbf{Detector} &\textbf{Input} & \textbf{Backbone} & \textbf{JHMDB} & \textbf{UCF101-24} \\ \Xhline{1.5pt}
        MOC*~\cite{moc} & \cmark & 7$\times$1 & DLA34 & 70.8 & 78.0 \\
        AVA*~\cite{ava} & \xmark & 20$\times$1 & I3D-VGG & 73.3 & 76.3 \\
        ACRN~\cite{acrn} & \xmark & 20$\times$1 & S3D-G & 77.9 &  - \\
        CA-RCNN~\cite{carcnn} & \cmark  & 32$\times$2 & R50-NL  & 79.2 & - \\
        YOWO~\cite{yowo} & \xmark & 16$\times$1 & 3DResNext-101 & 80.4 & 75.7 \\
        WOO~\cite{woo} & \xmark & 32$\times$2 & SF-R101-NL & 80.5 & - \\
        AIA~\cite{aia} & \cmark &  32$\times$1 & SF-R50-NL & - & 78.8 \\
        ACARN~\cite{acarn} & \cmark & 32$\times$1 & SF-R50  & - & \textbf{84.3} \\
        TubeR*~\cite{TubeR} & \xmark & 32$\times$2 & I3D  & - & 81.3 \\
        \hline
        STMixer & \xmark & 32$\times$2 & SF-R101-NL  & \textbf{86.7} & 83.7 \\ \Xhline{1.5pt}
    \end{tabular}
    }
    \caption{Comparison on JHMDB and UCF101-24. Methods marked with * use extra optical flow features.}
    \label{tab:jhmdb}
    \vspace{-3mm}
\end{table}

To verify the effectiveness of our STMixer, we further evaluate it on the JHMDB~\cite{jhmdb} and UCF101-24~\cite{ucf101} datasets. We report the frame-level mean average precision (frame-mAP) with an intersection-over-union (IoU) threshold of 0.5 in Table~\ref{tab:jhmdb}. Experimental results demonstrate that STMixer outperforms the current state-of-the-art methods with remarkable performance gain on JHMDB and achieves competitive results on UCF101-24. 

\subsection{Visualization}
We provide the visualization of sampling points and detection results of each ASAM module in order in Figure~\ref{fig:vis}. In the first few ASAM modules, the sampling points are quickly concentrated on the action performer and the actor localization accuracy gets improved rapidly. In the following ASAM modules, some of the sampling points get out of the human box and spread to the context. Benefiting from the gathered context and interaction information, the action recognition accuracy gets improved while localization accuracy is not compromised. In Figure~\ref{fig:vis}, we show clear improvements in three aspects: predicting missing classes (\textcolor{orange}{ride} in row 1),  improving the confidence score of correctly predicted classes (\textcolor[HTML]{23B051}{talk to} in row 2), and removing wrongly predicted classes (\textcolor{red}{watch} in row 3). To recognize an action ``ride'', we need to observe the person is in a car and someone is driving. For recognition of ``talk to", we need to know if a person is listening in the context, and for ``watch", we need to know if a person is in the actor's sight. The improvements are mostly related to these classes of interaction, which indicates our STMixer is capable of mining discriminative interaction information from the context.

\section{Conclusion and Future Work}

In this paper, we have presented a new one-stage sparse action detector in videos, termed STMixer. Our STMixer yields a simple, neat, and efficient end-to-end action detection framework. The core design of our STMixer is a set of learnable queries to decode all action instances in parallel. The decoding process is composed of an adaptive feature sampling module to identify important features from the entire spatiotemporal domain of video, and an adaptive feature mixing module to dynamically extract discriminative representations for action detection. Our STMixer achieves a new state-of-the-art performance on three challenging benchmarks of AVA, JHMDB, and UCF101-24 with less computational cost than previous end-to-end methods. We hope STMixer can serve as a strong baseline for future research on video action detectors.

One limitation of our STMixer is that the long-term query bank is implemented in an offline way where another STMixer without long-term query support is pre-trained for long-term query generation. We leave the design of an online query bank to future research and hope our STMixer is extended to extract long-form video information in an end-to-end manner.

\vspace{-3mm}
\paragraph {\bf Acknowledgements.} {\small This work is supported by National Key R$\&$D Program of China (No. 2022ZD0160900), National Natural Science Foundation of China (No. 62076119, No. 61921006, No. 62072232), Fundamental Research Funds for the Central Universities (No. 020214380091), and Collaborative Innovation Center of Novel Software Technology and Industrialization.}

{\small
\bibliographystyle{ieee_fullname}
\bibliography{egbib}
}

\clearpage
\section*{Appendix}

\section*{A. More Experimental Results}

\begin{figure*}[!h]
    \centering
    \includegraphics[width=16cm]{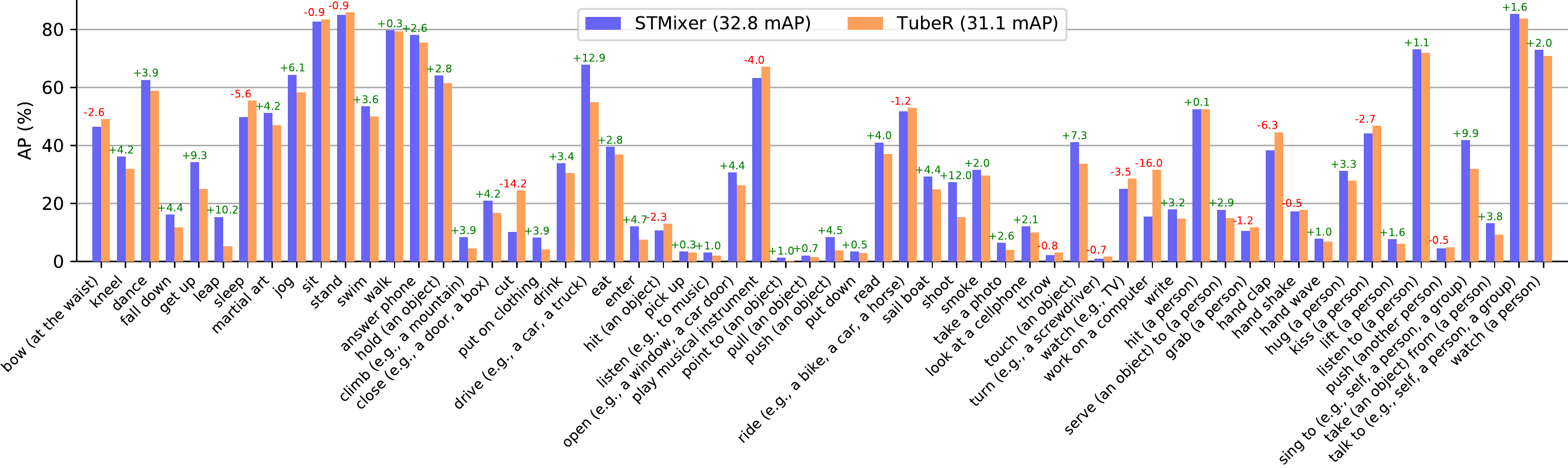}
    \scriptsize
    \vspace{-3mm}
    \caption{\textbf{AP of STMixer and TubeR on each action class of AVA v2.2.} We use the STMixer and TubeR models with the CSN-152 backbone for comparison. Both models do not use long-term features. When our STMixer has a higher AP, the difference is marked in \textcolor[HTML]{377E22}{green}, otherwise, it is marked in \textcolor{red}{red}. Our STMixer has a higher AP on most action classes than TubeR.}
    \label{fig:classAP}
\end{figure*}

\begin{figure*}[!h]
    \centering
    \includegraphics[width=15.5cm]{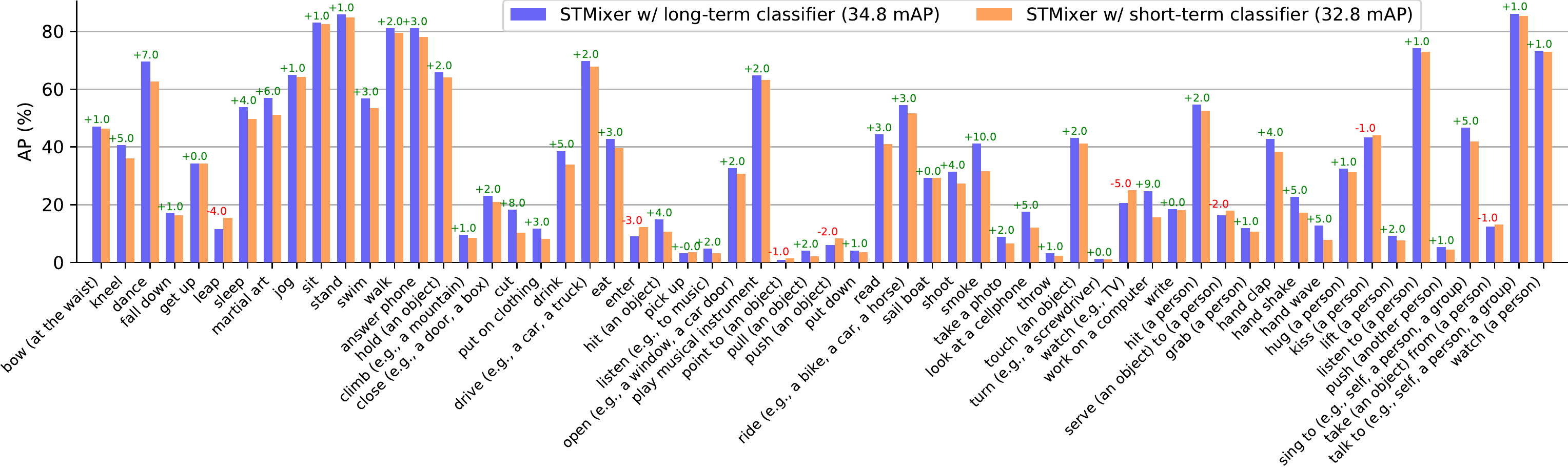}
    \scriptsize
    \vspace{-3mm}
    \caption{\textbf{AP of STMixer with a long-term or short-term classifier on each class of AVA v2.2.}  When STMixer with a long-term classifier has a higher AP, the difference is marked in \textcolor[HTML]{377E22}{green}, otherwise, it is marked in \textcolor{red}{red}.}
    \label{fig:lsclassAP}
\end{figure*}

\begin{figure*}[!h]
    \centering
    \includegraphics[width=16cm]{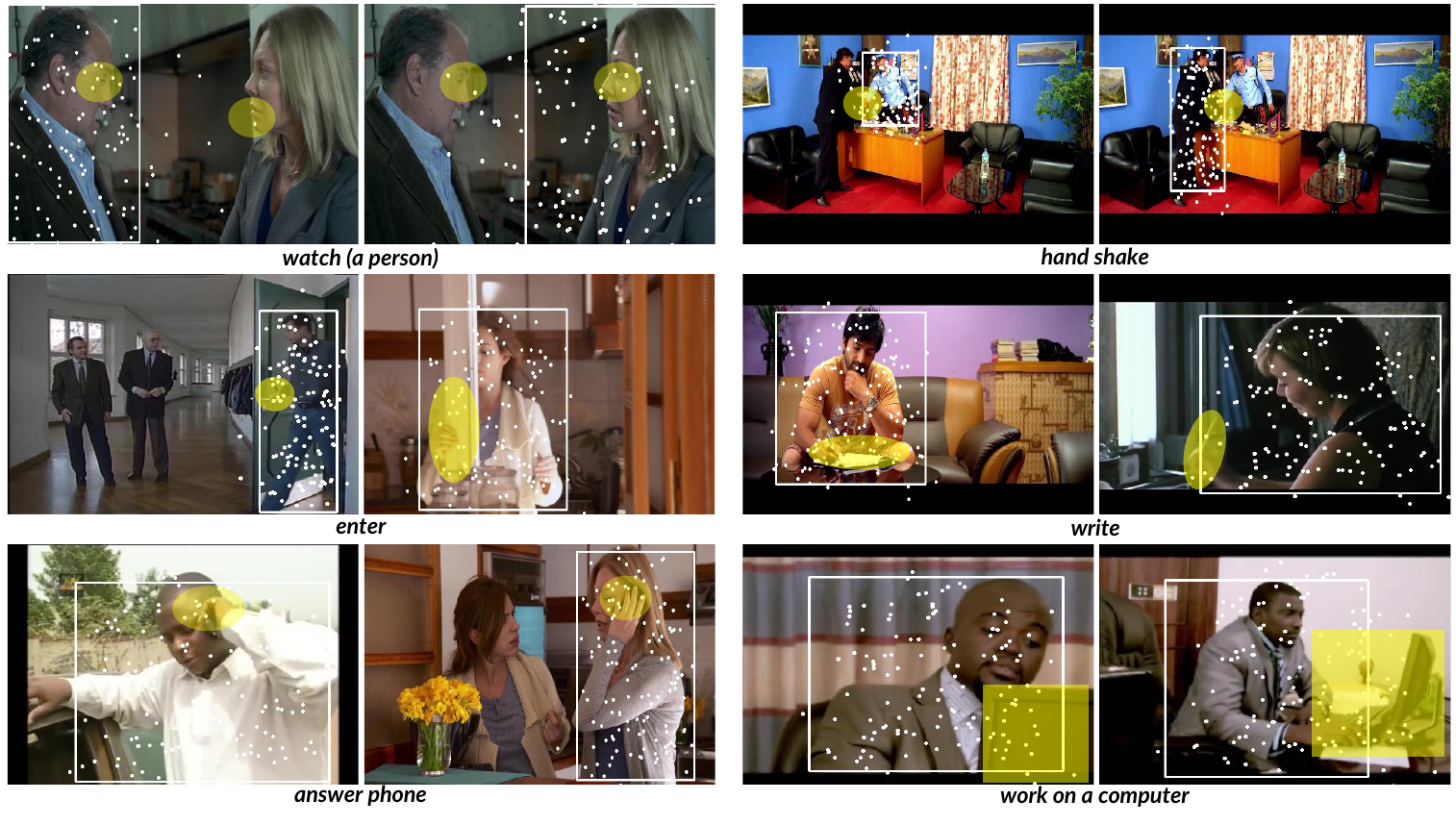}
    \scriptsize
    \vspace{-3mm}
    \caption{\textbf{Visualizations of sampled feature points for some action instances.} We show the sampled feature points of the last ASAM module. The yellow highlighted areas are considered to provide semantically relevant visual cues. }
    \label{fig:morevis}
\end{figure*}


\subsection*{A.1. AP on Each Action Class Comparison}
We provide detailed comparisons of the performance of STMixer and former state-of-the-art TubeR~\cite{TubeR} on each action class of AVA v2.2~\cite{ava} in Figure~\ref{fig:classAP}. For a fair comparison, both STMixer and TubeR models use the CSN-152 backbone~\cite{csn} and do not use long-term features. Out of all 60 classes, our method achieves higher AP on 47 classes, which makes the overall detection mAP of our STMixer higher by 1.7 than TubeR. We observe significant performance gaps on some interaction-related classes, such as interactions with objects (drive (\eg a car, a truck) \textcolor[HTML]{377E22}{$+12.9$}, shoot \textcolor[HTML]{377E22}{$+12.0$}), and interactions with other people (sing to (\eg self, a person, a group) \textcolor[HTML]{377E22}{$+9.9$}, take (an object) from (a person) \textcolor[HTML]{377E22}{$+3.8$}), which indicates our STMixer is more capable of modeling the relationship between the action performer and the surrounding objects and people. 

\subsection*{A.2. Impact of Long-term Classifier}
We provide the performance of STMixer with a short-term classifier or a long-term classifier on each action class of AVA v2.2 in Figure~\ref{fig:lsclassAP} to show the benefit of the long-term classifier. When using a long-term classifier, STMixer achieves better performance on the vast majority of action classes than using a short-term classifier, which demonstrates the importance of long-term information for action instance recognition. The experimental results also demonstrate that the action queries in our STMixer contain rich spatiotemporal information, and our design of long-term query bank and long-term query operation is effective. For action instances of some classes, the action performer sometimes interacts with objects or people appearing in other temporal segments.  For example, for an action instance of ``sing to (\eg self, a person, a group)", the singer and the listener are often in the different temporal segments of the video. We observe remarkable improvements on these classes (sing to (\eg self, a person, a group) \textcolor[HTML]{377E22}{$+5.0$}). STMixer with a long-term classifier can attend action queries to the long-term query bank for information of the listener. The long-term query bank and sampled features of the current video clip are complementary to each other and are both important for action detection~\cite{lfb}. This paper mainly focuses on the exploration of adaptive feature sampling from the feature space of the current video clip and adaptively feature mixing to enhance the representations, yet our simple design of long-term query back also yields good performance. 

\subsection*{A.3. Inference Speed Comparison}
\begin{table}[!h]
\centering
\large
\resizebox{\linewidth}{!}{
\begin{tabular}{lccccccc}
\Xhline{1.2pt}
\multirow{2}{*}{\textbf{Method}} &
  \multirow{2}{*}{\textbf{Extra Dect.}} &
  \multirow{2}{*}{\textbf{Input}} &
  \multirow{2}{*}{\textbf{Backbone}} &
  \multirow{2}{*}{\textbf{GFLOPs}} &
  \multicolumn{2}{c}{\textbf{mAP}} &
  \multirow{2}{*}{\textbf{FPS}} \\ \cline{6-7}
         &        &            &            &     & \textbf{v2.1} & \textbf{v2.2} &      \\ \Xhline{1.2pt}
SlowFast~\cite{slowfast} & \cmark & 32$\times$ 2 & SF-R101-NL & 365 & 28.2          & 29.0          & 5.8  \\
WOO~\cite{woo}      & \xmark & 32$\times$ 2 & SF-R101-NL & 252 & 28.0          & 28.3          & 6.9  \\
TubeR~\cite{TubeR}    & \xmark & 32$\times$ 2 & CSN-152    & 120 & 29.7          & 31.1          & 12.3 \\ \hline
STMixer      & \xmark & 32$\times$ 2 & SF-R101-NL & 135 & 29.8          & 30.1          & 11.6 \\
STMixer      & \xmark & 32$\times$ 2 & CSN-152    & 126 & 31.7          & 32.8          & 11.9 \\ \Xhline{1.2pt}
\end{tabular}
}
\caption{\textbf{Inference speed comparison on AVA dataset.} \cmark of column ``Extra Dect." denotes an extra human detector Faster RCNN-R101-FPN~\cite{fasterrcnn} is used. For a fair comparison, the resolution of input frames is set to 256$\times$256 for all models, and all models are tested on a GeForce RTX 3090 GPU.}
\label{tab:speed}
\end{table}

We compare the inference speed of our STMixer with former state-of-the-art methods in Table~\ref{tab:speed}. Methods like AIA~\cite{aia} and ACARN~\cite{acarn} follow the typical two-stage framework proposed by SlowFast~\cite{slowfast} but use a more complicated classification head for context modeling, so their complexity is higher than SlowFast. As AIA and ACARN do not report their complexity in their paper and these data are not available for us, we consider SlowFast as a lower bound of complexity for these methods here. For a fair comparison, long-term features are not used in all methods. As shown in Table~\ref{tab:speed}, because an extra human detection process is needed, two-stage methods like SlowFast have much lower inference speeds. Although training and inference are performed in an end-to-end manner, WOO~\cite{woo} still adopts a two-stage pipeline while TubeR and STMixer perform actor localization and action localization in one stage. This simplified pipeline makes inference speeds of TubeR and STMixer much higher. Compared to TubeR, our STMixer has 2.0 and 1.7 points higher mAP on AVA v2.1 and v2.2 respectively, while the inference speed of STMixer is comparable to TubeR (11.9FPS versus 12.3FPS). The training overhead of STMixer is also smaller than TubeR. STMixer converges within 10 epochs, while TubeR needs to be trained for 20 epochs despite using DETR~\cite{DETR} initialization. The adaptive sampling and adaptive mixing mechanism proposed in our STMixer makes it easier to cast the action queries to action instances.

\section*{B. More Visualizations}

As presented in the main paper, our proposed STMixer is a query-based framework for video action detection, which adaptively samples features from the spatiotemporal feature space without the restriction of human bounding boxes. In Figure~\ref{fig:morevis}, we provide more visualizations of sampled feature points for action instances. As shown in Figure~\ref{fig:morevis}, some of the sampling points go out of the human boxes and spread to other semantically related areas, which demonstrates the ability of our method to mine discriminative features from both the action performer itself and the surrounding context. 

\end{document}